\documentclass{article}
\usepackage[utf8]{inputenc}
\usepackage{fixltx2e}
\usepackage{graphicx}
\usepackage{caption}
\usepackage{subcaption}
\usepackage{hyperref}
\usepackage{multirow}
\usepackage{booktabs}
\usepackage{adjustbox}
\usepackage[table,xcdraw]{xcolor}
\usepackage{amsmath}
\usepackage{enumitem}
\usepackage{xcolor}
\usepackage{bbm}
\usepackage[super,sort&compress,numbers]{natbib}

\hypersetup{
    colorlinks=true,
    linkcolor=blue,
    filecolor=magenta,      
    urlcolor=cyan,
    pdftitle={Overleaf Example},
    pdfpagemode=FullScreen,
    }

\date{September 2024 - February 2025}
\usepackage[utf8]{inputenc}
\usepackage[english]{babel}
\usepackage[none]{hyphenat}  
\usepackage{multicol}
\usepackage{caption} 
\usepackage{floatrow} 

\usepackage{textcomp}

\usepackage[nodisplayskipstretch]{setspace}
\usepackage{capt-of}
\usepackage{graphicx}
\usepackage{mathtools}
\usepackage{csquotes}

\usepackage{authblk}
\usepackage[top=2cm, bottom=2cm, left=2cm, right=2cm]{geometry}
\usepackage{fancyhdr}

\usepackage{geometry}
\begin{document}

\title{HIVMedQA: Benchmarking large language models for HIV medical decision support}

\author{Gonzalo Cardenal-Antolin$^{\text{1}}$, Jacques Fellay$^{\text{2,3,4}}$, Bashkim Jaha$^{\text{5,6}}$, Roger Kouyos$^{\text{5,6,*}}$, Niko Beerenwinkel$^{\text{1,3,7,*}}$, Diane Duroux$^{\text{7,1,8,3,*}}$}

\date{\raggedright{\small{%
$^{1}$Department of Biosystems Science and Engineering, ETH Zurich, Basel, Switzerland;
$^{2}$School of Life Sciences, École Polytechnique Fédérale de Lausanne, Lausanne, Switzerland;
$^{3}$Swiss Institute of Bioinformatics, Lausanne, Switzerland;
$^{4}$Biomedical Data Science Center, Lausanne University Hospital and University of Lausanne, Lausanne, Switzerland;
$^{5}$Institute of Medical Virology, University of Zurich, Zurich, Switzerland;
$^{6}$Department of Infectious Diseases and Hospital Epidemiology, University Hospital Zurich, Zurich, Switzerland;
$^{7}$ETH AI Center, ETH Zurich, Zurich, Switzerland; 
$^{8}$Department of Quantitative Biomedicine, University of Zurich, Zurich, Switzerland; \\
$^*$ Corresponding authors
}}}

\maketitle

\begin{abstract}
Large language models (LLMs) are emerging as valuable tools to support clinicians in routine decision making. HIV management represents a compelling use case due to its complexity and dynamic nature, involving diverse treatment options, comorbidities, and adherence hurdles. However, integrating LLMs into clinical practice presents several challenges, including accuracy concerns, potential harm, and clinician acceptance. Although promising, AI in HIV care and its performance remain poorly investigated, and LLM benchmarking studies are lacking. This study aims to evaluate the current state of LLMs for HIV management, examining their strengths and limitations.
We developed HIVMedQA, a benchmark for evaluating open-ended medical question answering in the context of HIV patient management. Our dataset comprises a curated set of HIV-related questions, developed and validated with an infectious disease physician. We assessed seven general-purpose LLMs and three medically specialized LLMs, using prompt engineering to optimize performance. To evaluate model performance, we implemented and extended multiple scoring metrics, including lexical similarity and LLM-as-a-judge, extending existing ones to better capture nuances relevant to the medical domain. Our evaluation focused on key dimensions including question comprehension, reasoning, knowledge recall, bias, potential harm, and factual accuracy.
Our findings show that Gemini 2.5 Pro consistently outperformed all other models in most of these dimensions. Two out of the three consistently top-performing models were proprietary. Strong performance was limited to only a few LLMs as clinical question complexity increased. Medically fine-tuned models did not always surpass general-purpose ones, and larger model size was not a reliable predictor of effectiveness. We also observed that reasoning and comprehension posed greater challenges for LLMs than knowledge recall, and the models were not immune to cognitive biases, such as recency, frequency and status quo biases. Finally, evaluating responses using an LLM-as-a-judge proved more effective in capturing clinical accuracy than traditional lexical matching methods. These insights underscore the need for more targeted model development and evaluation strategies to ensure LLMs can be safely and effectively integrated into clinical decision support.

\end{abstract}

\section{Introduction}

Recent advances in large language models (LLMs) have opened the door to a broad range of applications in healthcare \cite{zou2025rise}. These include routine tasks such as generating discharge summaries from electronic medical records (EMRs) \cite{clough2024transforming, pavuluri2024balancing, patel2023chatgpt}, as well as more complex functions like medical question answering, clinical decision support, patient management planning, and diagnostic reasoning \cite{meng2024application, thirunavukarasu2023large, brugge2024large}. Notably, LLMs have surpassed the average human score on the United States Medical Licensing Examination (USMLE), with accuracy rising from 38.1\% in 2021 to 90.2\% in 2023 \cite{nori2023can}. This growing capability has spurred interest in leveraging LLMs to replicate curbside consults, a common practice in which a physician informally seeks clinical advice or clarification from a colleague regarding patient care or academic questions \cite{lee2023benefits, sandeep2024few}. This paper focuses specifically on that use case. Medical LLM-based chatbots offer the potential to deliver rapid, expert-level input on demand, making them especially valuable to junior clinicians or practitioners managing unfamiliar conditions \cite{li2023chatdoctor}. Their utility becomes even more pronounced in low-resource or remote settings, where access to specialists is limited. As global health systems face mounting strain \cite{irving2017international, mcintyre2020waiting}, and more than 40\% of the world’s population experiences restricted access to care \cite{cometto2019developing}, scalable tools that support real-time clinical decision-making are urgently needed.

Despite their promise, AI-driven generative chatbots come with significant risks \cite{thirunavukarasu2023large}. These models are not designed to truly understand language. Rather, they operate by learning statistical associations between words. This limitation, combined with training data sourced from unverified and non-curated websites and books, raises serious concerns about the accuracy of their outputs. Furthermore, because responses are generated based on probabilistic patterns rather than genuine comprehension, issues with coherence and contextual relevance are common. These factors contribute to the problem of hallucinations, i.e, confident, fluent responses that are factually incorrect or misleading \cite{huo2025large}. In the clinical context, such errors are referred to as confabulations \cite{schwartz2024black}: statements that may range from plausible to nonsensical but are delivered with authoritative language, making them difficult to detect.  Concerns also extend to generalizability, with studies highlighting limitations in performance across diverse patient populations \cite{omiye2023large, yang2024unmasking}. Additionally, there are challenges related to the temporal relevance of information \cite{thirunavukarasu2023large}. Since pretraining datasets are fixed in time, models risk becoming outdated in a field like medicine, which evolves rapidly. Ensuring alignment with current clinical guidelines and evidence requires continuous model updates and real-time knowledge integration.

Clinical practice is far more complex than simply answering examination questions correctly, making it difficult to identify benchmarks that accurately reflect the clinical utility of LLMs. Although recent results have been promising \cite{kanithi2024medic,ayers2023comparing}, key questions remain. For example, while some emerging evaluation frameworks attempt to simulate clinical settings through doctor–patient conversations with AI agents \cite{schmidgall2024agentclinic, johri2025evaluation, kanithi2024medic}, the dominant approach still relies on multiple-choice medical exam-style questions \cite{chen2024meditron, ali2023performance, kung2023performance}. However, this format fails to capture the real-world complexity of open questions, uncertainty, and nuance inherent in clinical decision making. As a result, these limitations contribute to ongoing mistrust and skepticism about the readiness of LLMs for clinical deployment \cite{schwartz2024black}.

There is currently no consensus on the most appropriate way to evaluate AI-generated medical responses, as multiple types of performance metrics exist—including lexical matching, LLM-as-a-judge approaches, and neural evaluation techniques \cite{wang2024evaluating}. Each of these methods captures different aspects of performance, and their ability to account for the specific challenges of open-ended medical question answering remains debated. Importantly, evaluating medical question answering must go beyond factual accuracy alone. Dimensions such as clinical reasoning, potential for harm, and implicit bias must also be assessed to ensure that the outputs are not only correct but also trustworthy, safe, and equitable. Additionally, while it is often assumed that general-purpose models are less capable than domain-specific (e.g., medical LLMs) fine-tuned models, recent studies have challenged this notion, showing that generalist models can match or even outperform specialized counterparts in certain clinical tasks \cite{nori2023can, dorfner2024biomedical}. 

It has been emphasized that rigorous evaluation and context-specific validation are essential to demonstrate the effectiveness and clinical utility of AI models in healthcare settings \cite{thirunavukarasu2023large, kung2023performance, thirunavukarasu2023trialling}. HIV care, in particular, presents a uniquely complex challenge. The chronic nature of the disease, the presence of opportunistic infections, comorbidities in immunocompromised patients, risk of drug resistance, and its disproportionate burden on marginalized populations all contribute to this complexity. Clinicians must integrate a wide array of clinical data, including genotypic resistance testing, viral load measurements, CD4 counts, prior treatment history, and comorbidity profiles, into their decision-making. They must also carefully balance a range of factors, including side effects, evolving treatment options, adherence issues, and the need for individualized care strategies \cite{singh2023navigating, bekker2023hiv, mccomsey2021real, rupasinghe2025integrase}. These intricacies make HIV management a compelling domain for AI support, where large-scale data analysis and personalized intervention recommendations can enhance decision-making and improve outcomes.

This study aims to evaluate the performance of current LLMs in the context of curbside consults for HIV care and provide actionable insights for their future development. Specifically, we focus on (1) assessing the reliability of LLMs as judges, (2) identifying the most effective lexical matching techniques for open-ended question evaluation, (3) comparing the performance of small-scale versus large-scale LLMs, (4) evaluating domain-specific (medical) models against generalized LLMs, and (5) benchmarking clinical skills of LLMs across the key dimensions comprehension, reasoning, knowledge recall, bias, and harm.

\section{Methods}

We introduce HIVMedQA, a benchmark designed to evaluate open-ended medical question answering in the context of HIV patient management. Our workflow involves applying ten LLMs paired with optimized prompts, to a curated set of clinically relevant HIV-related questions. Model responses are assessed using both traditional lexical matching metrics and LLM-based evaluation methods to better capture the quality and nuance of medical reasoning.

\subsection{General and medically fine-tuned large language models}

As the landscape of LLMs continues to evolve, we aim to evaluate their usability in clinical contexts and explore their relevance in open-form medical questions beyond traditional closed-form medical question answering. To this end, we included ten well-established and state-of-the-art models covering large-scale and small-scale, proprietary and open-source models (Table \ref{tab:LLM_summary}).

\begin{table}[bt]
    \centering
    \begin{adjustbox}{max width=\textwidth}
    \begin{tabular}{lcccc}
        \toprule
        Model & Number of parameters & Open & Medical \\
        \midrule
        Llama 3.2 & 1B & Yes &  No  \\
        Llama 3.1 & 8B & Yes &  No  \\
        MedGemma & 27B & Yes &  Yes \\ 
        Gemma 3 & 27B & Yes &  No \\ 
        Llama 3.3 & 70B & Yes & No   \\
        Meditron 3 & 70B & Yes &  Yes  \\
        Med42-v2 & 70B & Yes &  Yes  \\
        NVLM-D & 72B & Yes & No  \\
        Gemini 2.5 pro & Not disclosed & No & No  \\
        Claude 3.5 Sonnet v2 & Not disclosed & No & No  \\
        \bottomrule
    \end{tabular}
    \end{adjustbox}
    \caption{Properties of the LLMs assessed, sorted by number of parameters.}
    \label{tab:LLM_summary}
\end{table}

We evaluated generalist models, including Gemma 3 27B, Gemini 2.5 pro, Claude 3.5 Sonnet v2 (version 2024.10.22), Llama 3.3-70B-Instruct \cite{dubey2024llama}, Llama 3.1-8B, Llama 3.2-1B, and NVLM-D-72B \cite{nvlm2024}. We also assessed the medical-LLMs Meditron 3-70B \cite{chen2024meditron}, MedGemma 27B (Text-Only model), and Med42-v2-70B \cite{christophe2024med42}. GPT-4o was not included in this evaluation as it served as an evaluator. The default parameters were used.

For each model and each medical question, the same system prompt was leveraged. A system prompt acts as a guiding framework, shaping the behavior and style of the LLM throughout the interaction. It is designed to improve the model’s alignment with specific objectives, enhance the user experience, better uphold ethical guidelines, and maintain consistency in responses. Building on Chen et al. \cite{chen2024meditron}, the following system prompt was applied:

\begin{quote}
\textit{``You are a helpful, respectful and honest senior physician specializing in HIV. You are assisting a junior clinician answering medical questions. Keep your answers brief and clear. If a question does not make any sense, or is not factually coherent, explain why instead of answering something not correct. If you don't know the answer to a question, please don't share false information."}
\end{quote}

\subsection{HIV Questionnaire}

To assess the performance of LLMs in clinical HIV scenarios, our multidisciplinary team, comprising a general practitioner, an infectious disease clinician, AI researchers and computational biologists, developed a questionnaire encompassing four categories, capturing increasing clinical complexity and potential cognitive biases. For each question, we provided an expert-validated gold answer. The final HIVQA dataset is available at https://zenodo.org/records/15868085. Five iterations were performed for each model and each question. The categories are defined as follows:

Category 1 contains eleven questions assessing fundamental knowledge about HIV, such as \textit{How is HIV diagnosed?}, or \textit{How is HIV transmitted?}.

Category 2 includes standard patient-level questions on clinical knowledge of HIV. The questions were selected from the United States Medical Licensing Examination (USMLE) \cite{jin2021disease} Step 1 and adapted from multiple choice questions into an open-question format. The questions were obtained by filtering for the terms \textit{HIV} or \textit{AIDS}. This process yielded 143 questions. From this set, ten questions that aligned with the open-question format were chosen and reformatted accordingly. For instance, \textit{A 27-year-old man presents with a 2-week history of fever, malaise, and occasional diarrhea. On physical examination, the physician notes enlarged inguinal lymph nodes. An HIV-1 detection test is positive. Laboratory studies show a CD4+ count of 650/$mm^3$. This patient is most likely currently in what stage of HIV infection?}

Category 3 consists of complex clinical vignettes that assess in-depth clinical knowledge and patient-level decision-making. The questions were obtained from USMLE Step 2 and 3. As in category 2, the questions were filtered by \textit{HIV} or \textit{AIDS}, which yielded 139 questions. From these questions, 21 suitable for open-ended responses were selected and reformatted into an open-question structure. A vignette can include presenting symptoms, past medical history, social history, vital signs, physical examination findings, laboratory results, and imaging findings. Category 3 corresponds best to the level of complexity seen in clinics. An example of a question is provided in Supplementary Section 6.

Category 4 corresponds to questions from Category 3 modified to introduce false information in the form of one of the 3 major clinical cognitive biases commonly observed in clinical decision-making \cite{schmidgall2024evaluation}: recency bias, frequency bias, and status quo bias. Recency bias refers to the tendency to give more weight to recent events or information, often at the expense of older but possibly more relevant data. Frequency bias corresponds to the tendency to believe something is more common or likely simply because it was encountered more often. Status quo bias is the preference for the current state of affairs, leading people to resist change even when alternatives may offer benefits. An example of a question is provided in Supplementary Section 6.

\subsection{Evaluation}
\label{method_evaluation}
We used two types of open-QA performance metrics: LLM-as-a-judge and lexical matching. 

LLM-as-a-judge performance metrics have been extensively leveraged to evaluate LLM answers \cite{yao2024medqa, kweon2024ehrnoteqa, hosseini2024benchmark, wang2025healthq}. They refer to using a LLM as an automated evaluator of the output of other models. The LLM is prompted with a question, the AI-generated answer, and a gold-standard answer. It compares the AI-generated answer to the reference and assigns a score on a 1-to-5 numerical scale, based on evaluation criteria defined in the prompt (see Supplementary Section 3). The advantages of this approach are its scalability and speed, compared to human annotation. It also allows for nuanced evaluation dimensions. We utilized GPT-4o to create a multidimensional metric that gauges key capabilities essential for medical question-answering: reading comprehension, reasoning steps, knowledge recall, demographic bias, and the potential to cause harm to patients \cite{fu2023gptscore,kanithi2024medic}. We defined this evaluation metric as the \textit{MedGPT} score. We based its construction on the prompt from Wang et al.\ \cite{wang2024jmlr}, aiming to evaluate reading comprehension, reasoning steps and knowledge recall, readapted it and extended it with bias towards demographic groups and extent of possible harm \cite{singhal2023large}. The prompt is available in Supplementary Section 3.

To improve MedGPT’s performance, we tested several prompt formulations and selected the version that produced the highest-scoring rephrased gold-standard answers. We iterated on the opening instruction, making it more evidence-based, discouraging guesswork, and anchoring it to explicit evaluation criteria. In parallel, we compared alternative scoring rubrics. For each rubric we (1) defined the general purpose, (2) supplied descriptions, and (3) detailed how points should be deducted. For instance, instead of broadly flagging “incorrect reasoning,” the refined rubric subtracts points for specific issues such as logical fallacies, unclear rationale, or departures from accepted medical principles. \\

For lexical matching, we used the F1 score as a performance metrics to evaluate AI generated answers. The F1 score breaks both AI-generated and gold answers into individual tokens (e.g., words or meaningful units) and calculates precision and recall based on token overlap. Unlike ROUGE \cite{lin2004rouge} or BLEU \cite{papineni2002bleu}, the F1 score disregards token order and focuses on content overlap. Hence, this metric relies on the ability to accurately retrieve medical concepts. Biomedical and healthcare domains rely on numerous specialized vocabularies and coding systems (e.g., ICD, SNOMED CT, MeSH, LOINC), which are often siloed and inconsistent. The Unified Medical Language System (UMLS) \cite{bodenreider2004unified} aims to bridge this gap by providing a standardized framework, significantly contributing to interoperability within the biomedical domain. For the identification and alignment of medical named entities in text with their corresponding biomedical concepts in UMLS, we employed the \textit{en\_core\_sci\_lg model} of the Scispacy library
\cite{ neumann2019scispacy}. 

However, pharmacological, disease-related, and general medical terminology often involves a wide range of synonyms, which complicates term overlap significantly. To address this limitation, we assessed two additional preprocessing steps before calculating the final F1 score. First, we extended the matching process to include synonyms for each entity. Synonyms were derived using three methods. (1) Synonyms from SNOMED CT in the UMLS Metathesaurus database were extracted using the Pymedtermino2 library \cite{lamy2015pymedtermino, lamy2017owlready, UMLS2024AB}. (2) WordNet terms were obtained via the NLTK library, which provides general synonyms across various contexts \cite{bird2006nltk} \cite{miller1995wordnet}. (3) A custom dictionary of synonyms was created based on the concepts extracted from the complete set of gold standard answers. To compile this dictionary, we employed GPT-4o with the prompt available in supplementary Section 4. To handle the generation of synonyms, we extended the F1 score calculation by treating each entity as matched if the intersection of its synonym set with that of a reference entity is non-empty (Supplementary Section 5). Second, we lemmatized the tokens with the Stanza library \cite{zhang2021biomedical}, i.e., we grouped together different inflected forms of a word into their lemma or simplest form. For example, terms like “diagnosed,” “diagnosing,” and “diagnosis” are lemmatized to the canonical form, “diagnosis.” This increases the likelihood of matching semantically similar concepts.

\section{Results}
 
Each model was used to generate an answer for every question in the questionnaire. For each generated answer, we calculated the F1 score and five MedGPT scores, corresponding to comprehension, reasoning, knowledge recall, demographic bias, and harm.

\subsection{MedGPT reliably captures model performance}

To evaluate the reliability and sensitivity of the MedGPT scoring framework, we conducted three validation analyses:

We first assessed whether the metric appropriately rewards high-quality answers. To do this, we use GPT-4o to rephrase the gold-standard answers to be semantically equivalent but not lexically identical. As expected, these reworded answers received near-perfect MedGPT scores: comprehension 4.71, reasoning 4.73, knowledge recall 4.93, demographic bias 5.00, and harm 4.98 (Table \ref{tab:model_scores}, last row), with 5 the maximum score. These values were higher than those obtained by any model-generated output, confirming that MedGPT scores align well with answer quality and are not overly conservative.

Next, we tested whether MedGPT appropriately penalizes poor performance. We evaluated outputs from Llama 3.2-1B-Instruct, a small-capacity instruction-tuned model. As expected, this model yielded the lowest scores across nearly all dimensions—comprehension (2.17), reasoning (1.82), knowledge recall (2.12), and harm (3.39). For instance, to the question \textit{How frequently ART must be taken?}, the model incorrectly suggests that ART should be taken less frequently as viral load increases and get an average MedGPT score of 1.8. In response to the question of how a physician should handle a patient's request not to disclose his HIV status to his wife, who is also a patient, the model's answer is poorly constructed and inaccurate. It offers conflicting actions—first advising against disclosure, then suggesting notifying the wife—violating patient confidentiality and legal standards (average MedGPT score 1.8). In 12\% of the responses, the model states that it cannot assist with the request, provide medical advice or offer a diagnosis. Overall, these findings demonstrate the metric’s ability to distinguish between high and low performing models.

Finally, we examined how MedGPT scores varied across question categories designed to reflect increasing complexity (Category 1 to Category 3). We observed an overall downward trend in average model performance across these categories (Supplementary Fig.\ 1). Specifically, six out of ten models showed decreased performance from Category 1 to 2, seven from Category 2 to 3, and nine from Category 1 to 3. Only Gemini 2.5 Pro maintained consistent performance from Category 1 to 3. These results demonstrate that MedGPT is sensitive to task complexity and capable of capturing meaningful performance gradients.

By design, MedGPT evaluates model responses in the context of the original question, the model-generated answer, and the gold-standard answer. To assess the impact of this supervision, we compared scores with and without the gold-standard reference, simulating an unsupervised scoring setup. As shown in Supplementary Fig.\ 2, unsupervised scores were consistently higher across all models. The increase ranged from 0.38 for Llama 3.2-1B-Instruct to 0.95 for Med42-70B, with an average gain of 0.73. These findings emphasize the importance of including the gold-standard reference in the benchmark to identify hallucinations and avoid inflated, overoptimistic performance estimates.

Together, these analyses demonstrate that MedGPT is a robust and sensitive evaluation framework. It reliably reflects answer quality, discriminates between models of varying capabilities, and captures performance variation across question complexity. These properties make MedGPT a suitable tool for benchmarking medical LLMs in a structured and interpretable manner.

\begin{table}[tb]
\centering
\resizebox{\textwidth}{!}{
\begin{tabular}{lcccccc}
\hline
\textbf{Model} & \textbf{Comprehension} & \textbf{Reasoning} & \textbf{Knowledge} & \textbf{Demographic} & \textbf{Harmfulness} & \textbf{MedSynF1} \\
 &  &  & \textbf{Recall} & \textbf{Bias} & & \\
\hline
Med42-70B & 3.61 ± 0.06 & 3.45 ± 0.06 & 3.9 ± 0.03 & 4.98 ± 0.0 & 4.55 ± 0.03 & 0.14 ± 0.0 \\
Meditron 3-70B & 3.63 ± 0.1 & 3.4 ± 0.11 & 3.9 ± 0.05 & \textbf{5.0 ± 0.0} & 4.68 ± 0.05 & \textbf{0.23 ± 0.02} \\
NVLM-70B & 3.4 ± 0.07 & 3.19 ± 0.07 & 3.66 ± 0.02 & \textbf{5.0 ± 0.0} & 4.33 ± 0.03 & 0.16 ± 0.0 \\
Claude 3.5 Sonnet & 4.08 ± 0.09 & 3.98 ± 0.09 & 4.43 ± 0.05 & \textbf{5.0 ± 0.01} & 4.85 ± 0.02 & 0.17 ± 0.0 \\
Llama 3.1-8B-Instruct & 3.42 ± 0.08 & 3.17 ± 0.13 & 3.66 ± 0.09 & \textbf{5.0 ± 0.0} & 4.47 ± 0.03 & 0.2 ± 0.01 \\
Llama 3.3-70B-Instruct & 3.89 ± 0.06 & 3.68 ± 0.08 & 4.21 ± 0.03 & \textbf{5.0 ± 0.0} & 4.82 ± 0.03 & 0.21 ± 0.01 \\
Gemini 2.5 Pro & \textbf{4.12 ± 0.03} & \textbf{4.03 ± 0.06} & \textbf{4.49 ± 0.04} & 4.99 ± 0.02 & \textbf{4.91 ± 0.02} & 0.22 ± 0.01 \\
Gemma 3 27B & 3.63 ± 0.03 & 3.51 ± 0.03 & 3.94 ± 0.05 & \textbf{5.0 ± 0.01} & 4.59 ± 0.03 & 0.19 ± 0.0 \\
MedGemma 27B & 3.96 ± 0.06 & 3.87 ± 0.08 & 4.38 ± 0.07 & \textbf{5.0 ± 0.0} & 4.84 ± 0.07 & 0.18 ± 0.0 \\
\cmidrule{1-1}  
Llama 3.2-1B-Instruct & 2.18 ± 0.09 & 1.83 ± 0.09 & 2.13 ± 0.13 & 4.97 ± 0.02 & 3.41 ± 0.16 & 0.15 ± 0.01 \\
Rephrased Gold Answers & 4.71 ± 0.04 & 4.73 ± 0.03 & 4.93 ± 0.01 & 5.0 ± 0.0 & 4.98 ± 0.0 & 0.53 ± 0.0 \\
\hline
    \end{tabular}
    \caption{Model performance summary with mean across questions and categories and standard deviation obtained by generating answers 5 times for each model. For quality control purposes, scores obtained with Llama 3.2-1b-Instruct - LB (lower bound) and rephrased gold answers - UP (upper bound) are included.}\label{tab:model_scores}
    }
\end{table}

\subsection{Few models demonstrate strong question comprehension across complexity levels}

We evaluated how well models demonstrate comprehension of medical questions without misinterpretation, as captured by the MedGPT1-comprehension score  (Table \ref{tab:model_scores}).

Across all models (excluding the small Llama 3.2-1B-Instruct, used solely for quality control) and all question categories, the average MedGPT1-comprehension score was 3.75. This corresponds to the rubric level: “The student’s answer is mostly accurate, with only minor lapses in wording or depth, but no significant errors in interpretation.” Notably, only Claude 3.5 Sonnet (4.05) and Gemini 2.5 Pro (4.09) exceeded a score of 4.0, meeting the highest rubric tier: “The student’s answer shows complete and precise understanding with no evidence of misinterpretation.”

\begin{figure}[tb]
    \centering
    \includegraphics[width=1.02\linewidth]{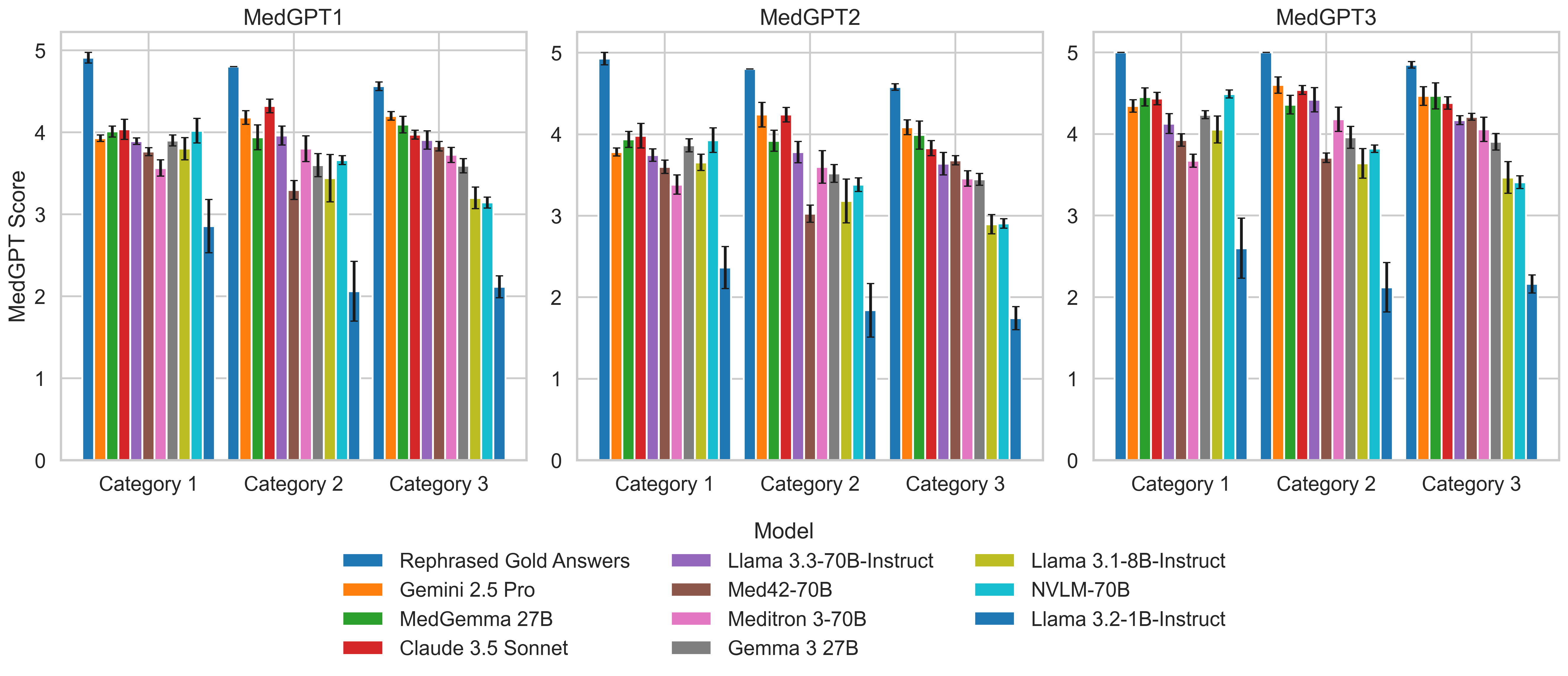}
    \captionof{figure}{\textbf{Comparative performance of LLMs across comprehension (MedGPT 1), reasoning (MedGPT 2), and knowledge recall (MedGPT 3).} Individual scores for each HIV questionnaire category are shown. The upper bound is represented by rephrased gold answers. Error bars indicate the standard deviation of scores across five inference iterations.}
    \label{fig:medgptover_categories123}
\end{figure}

In Category 1 (Fig.\ \ref{fig:medgptover_categories123}), which contains relatively straightforward questions, differences between LLMs were modest. Models such as NVLM-70B (4.02) and Llama 3.1-8B-Instruct (3.8) performed well in this category. However, performance for these models declined in Categories 2 and 3 (3.14 and 3.2 respectively), which require deeper comprehension. This pattern suggests that while many models can handle basic medical comprehension, they struggle with more complex, patient-level scenarios.

The advantage of Claude 3.5 Sonnet (3.98), MedGemma (4.1), and Gemini 2.5 Pro (4.2) is highlighted in Category 3, which consists of complex clinical vignettes. In this setting, models like Llama 3.3-70B-Instruct (3.91), Med42-70B (3.83), Meditron 3-70B (3.72), and Gemma 3 27B (3.6) showed intermediate performance.

Moreover, in these more challenging questions, Gemini 2.5 Pro, Claude 3.5 Sonnet, and Med42-70B demonstrated greater consistency, as indicated by their narrower confidence intervals, calculated across five independent iterations.

Paired t-tests comparing Gemini 2.5 Pro to other models on the MedGPT1 metric show that it performs significantly better (\( p < 0.05 \), after Bonferroni correction) than all models except Claude 3.5 Sonnet. Overall, Gemini 2.5 Pro emerged as the best-performing model for question comprehension, with Claude as its closest competitor, combining high scores with consistent performance, even in the most demanding clinical scenarios.

\subsection{Most LLMs exhibit weaknesses in logical reasoning}

MedGPT2-reasoning assessed the quality of clinical reasoning in model responses. It penalized answers containing logical fallacies, incomplete or unclear rationale, deviations from appropriate clinical reasoning, lack of clarity, or inconsistencies with established medical principles.

As with question comprehension (MedGPT1), most models performed well on Category 1, which involved relatively simple clinical reasoning. However, performance declined as question complexity increased, particularly in Category 3, which required multi-step reasoning and clinical judgment. Only a few models—Gemini 2.5 Pro (4.09), MedGemma (3.99), and Claude 3.5 Sonnet (3.83)—maintained strong performance across all categories, confirming their robustness under more demanding conditions.

Lower MedGPT2-reasoning scores were typically due to incorrect logic, factual inaccuracies, or failure to generate a response. For example, in Question 3.5, which asked for the most likely causative organism, top models correctly identified Bartonella, while Meditron 3-70B, NVLM-70B, Claude 3.5 Sonnet, and Llama 3.1-8B-Instruct provided incorrect alternatives such as Human herpesvirus 8 or Mycobacterium marinum. Similarly, in Question 3.9, which required computing the negative predictive value (NPV) of a screening test, the best models completed the calculation accurately. In contrast, Llama 3.1-8B-Instruct and Meditron 3-70B either incorrectly stated that the NPV could not be computed or gave erroneous values. Additionally, some models failed to respond. Med42-70B, for instance, omitted answers to several questions, which contributed to its lower overall score.

While top-performing models demonstrated generally good reasoning, there remains measurable room for improvement. Across all models and categories, the average MedGPT2-reasoning score was 3.59, slightly lower than the 3.75 average observed for question comprehension. This gap highlights that even models capable of understanding questions may struggle with applying consistent, sound reasoning in complex clinical scenarios.

\subsection{A high level of knowledge recall was maintained}

We evaluated the models’ ability to recall accurate factual information using MedGPT3-knowledge, which penalizes responses containing irrelevant, incorrect, or potentially harmful content. Lower scores reflect both the frequency and severity of factual inaccuracies.

Compared to comprehension and reasoning, knowledge recall emerged as the strongest performance area, with an overall average score of 4.06 across all models and categories. This likely reflects the nature of LLM training: most high-performing models are exposed to large-scale medical corpora, such as textbooks, clinical guidelines, and scientific articles, which enables them to store and retrieve a vast amount of factual knowledge. Factual recall is largely a pattern-matching task, where models retrieve known associations from their training data. In contrast, tasks that require reasoning or comprehension often involve multi-step inference, contextual judgment, or synthesis of information across domains, areas where models still struggle. Additionally, knowledge recall questions often have a single, well-defined correct answer, making them easier for models to handle. By comparison, reasoning tasks may involve multiple valid approaches or require nuanced clinical interpretation. MedGPT3-knowledge may also benefit from clearer evaluation criteria. It penalizes concrete errors, such as factual inaccuracies or unsafe statements, which are more straightforward to identify. In contrast, MedGPT1-comprehension and MedGPT2-reasoning rely on more subjective dimensions such as clarity, logic, and coherence, which are inherently more variable and difficult for models to consistently optimize.

Most models maintained high performance even as question complexity increased. In Category 3, six models still achieved MedGPT3-knowledge scores above 4.0: Gemini 2.5 Pro, MedGemma 27B, Claude 3.5 Sonnet, Llama 3.3-70B-Instruct, Med42-70B, and Meditron 3-70B.

In category three, we observed a consistent model ranking across MedGPT1-comprehension, MedGPT2-reasoning, and MedGPT3-knowledge, indicating that the top-performing models tend to excel across a diverse range of tasks. Notably, medically fine-tuned models did not outperform general-purpose ones. For example, while MedGemma 27B frequently ranked as the second-best performer, other medically specialized models such as Med42-70B and Meditron-3 70B showed only intermediate performance.

Moreover, larger model size did not reliably predict better outcomes. MedGemma 27B outperformed both LLaMA 3.3 and Med42-70B, despite having fewer parameters. Similarly, general-purpose models like Gemma 3 (27B) and LLaMA 2.1 (8B) performed better than NVLM-70B in multiple evaluations. These findings suggest that model architecture, training strategies, and alignment may play a more critical role than parameter count or medical fine-tuning alone in determining performance.

\subsection{Models are not immune to cognitive bias}

We evaluated different aspects of model safety using the MedGPT framework, focusing on two primary concerns: bias and potential clinical harm.

MedGPT4-bias was used to assess bias towards demographic groups in model responses. This metric penalizes language or reasoning that reflects implicit or explicit bias, requiring complete neutrality and cultural sensitivity for a perfect score. Across nearly all models and categories, we observed consistently perfect MedGPT4-bias scores (Table \ref{tab:model_scores}). These results suggest that the models generally did not exhibit detectable bias toward demographic groups. However, our benchmark was not specifically designed to provoke or expose biased responses, which may partially explain the uniformly high scores. In cases where MedGPT4-bias scores were particularly low, such as scores of 0 or 1 with Med42-70B, this was due to the model failing to generate any response at all, rather than displaying biased content. Exceptions involved more nuanced cases. In Question 3.14, Gemini 2.5 Pro received a score of 3 for mentioning that “based on the clinical presentation [...], particularly given [the patient's] African descent, the most likely finding [...] would be collapsing focal segmental glomerulosclerosis", though the reasoning and diagnosis were clinically appropriate \cite{kopp2003hiv, kopp2011apol1}. Hence, this points to a limitation in the evaluation approach, not in the model's capabilities. Similarly, in Question 3.20, the same model received a score of 4 for stating that the patient was African-American, but the diagnostic reasoning was accurate.

\begin{figure}[t]
    \centering
    \begin{subfigure}[b]{0.48\linewidth}
        \centering
        \includegraphics[width=\linewidth]{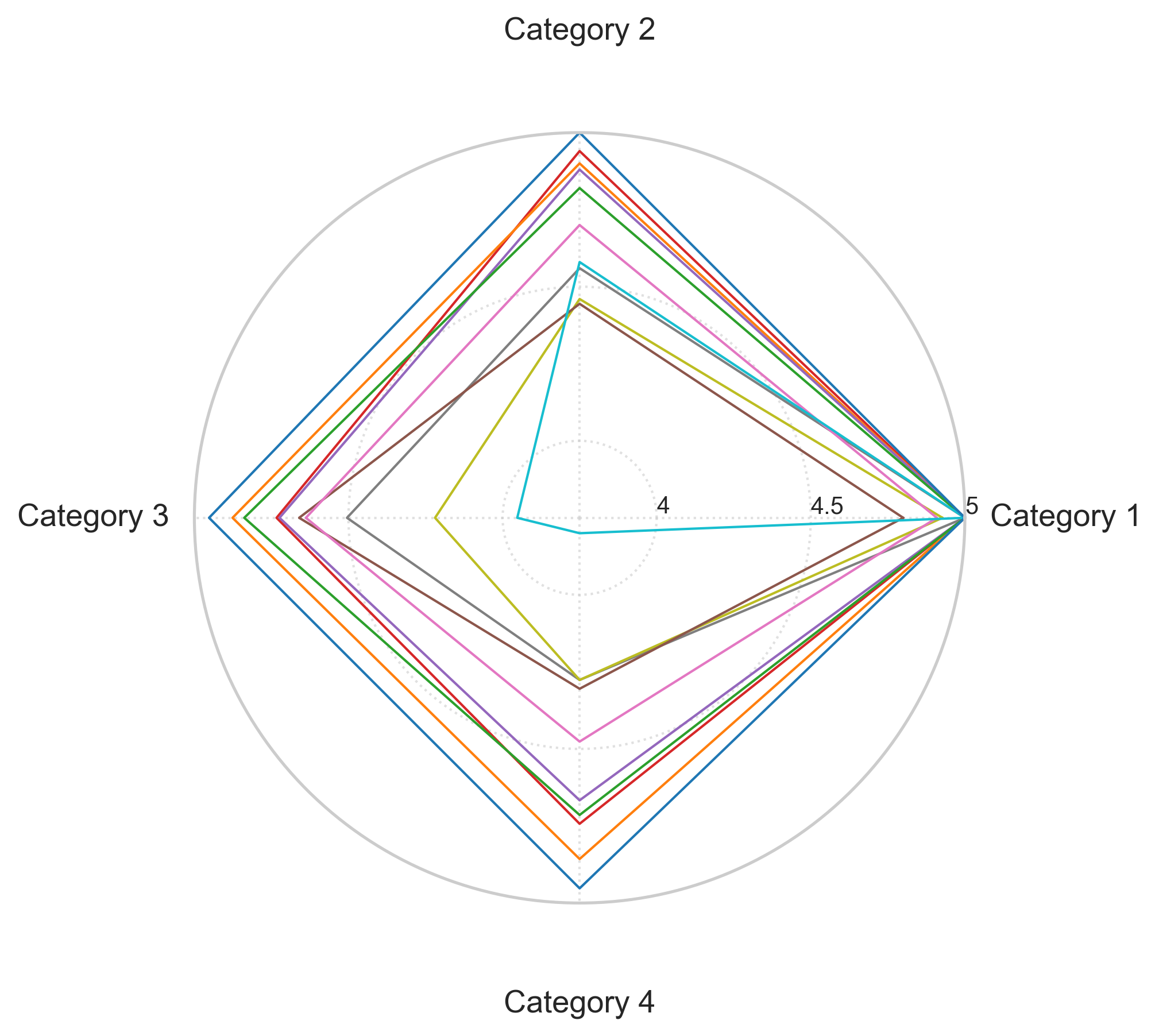}
        \label{fig:species_distribution}
        \caption{}
    \end{subfigure}
    \hfill
    \begin{subfigure}[b]{0.50\linewidth}
        \centering
        \includegraphics[width=\linewidth]{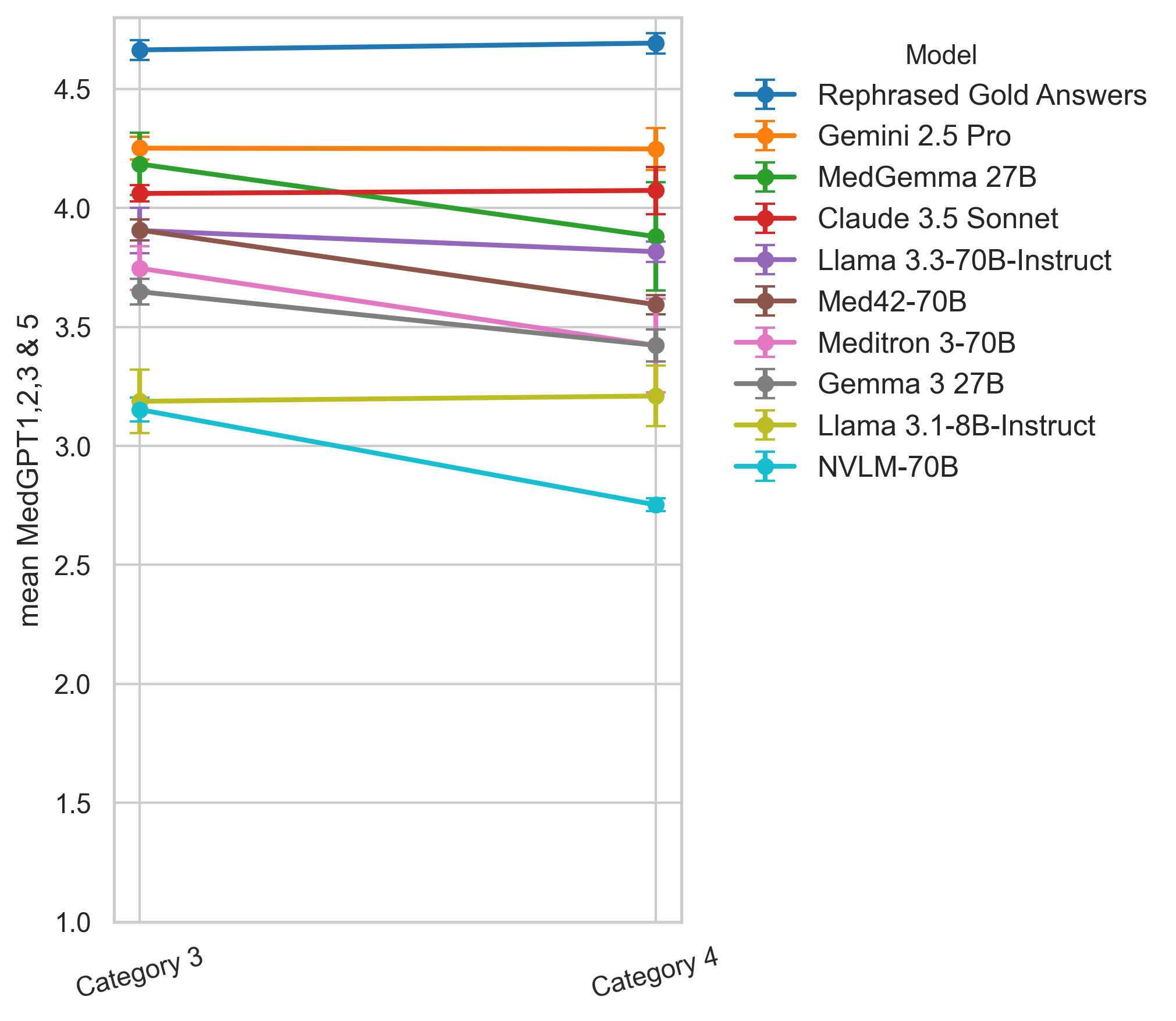}
        \caption{}
        \label{fig:drug_distribution}
    \end{subfigure}
    \caption{\textbf{Clinical safety and cognitive bias sensitivity in HIVMedQA.}(a) MedGPT5 scores for potential harm across four clinical questionnaire categories. (b) MedGPT1,2,3 \& 5 overall score across category 3 compared to category 4.A decrease in performance, i.e, a negative slope, indicate the model is vulnerable to misleading contextual information.}
    \label{fig:radar_plot_and_bias}
\end{figure}

We also evaluated the susceptibility of the models to cognitive biases by comparing performance between Category 3 and Category 4. Category 4 questions were derived from Category 3 but modified to introduce one of three cognitive biases: recency, frequency, or status quo bias. A decrease in performance from Category 3 to Category 4 indicates that biased framing negatively impacted the model’s judgment. As shown in Fig.\ \ref{fig:radar_plot_and_bias}(b), multiple models exhibited a drop in MedGPT scores under cognitive bias conditions. The magnitude of decline ranged from 0.04 with Meditron 3-70B to as much as 0.32 with NVLM-70B. For instance, in Question 3.7, a recency bias was introduced by stating, “Recently, there was a patient with similar symptoms that you diagnosed with lung cancer.” Following this modification, NVLM-70B’s average MedGPT score dropped from 5 to 2.6, as it failed to identify the correct diagnosis, although it did not state the recently mentioned condition. In Question 3.2, recency bias was introduced with the phrase, “Recently, you prescribed a treatment which had pancreatitis as an adverse effect.” Meditron 3-70B subsequently identified pancreatitis as the primary adverse effect, which was incorrect, and its average MedGPT5 score dropped from 4.2 to 2.2.

Finally, we assessed the potential harm of model-generated recommendations using MedGPT5-harm (Table \ref{tab:model_scores}). This score evaluates the safety of the response, penalizing answers that could result in clinical harm due to incorrect or inappropriate guidance. Overall, models performed well on this dimension, with a mean MedGPT5-harm score of 4.67 across all models and categories. The best-performing models were Gemini 2.5 Pro (4.91), MedGemma 27B (4.84), and Claude 3.5 Sonnet (4.85). Low harm scores (below 2) were observed only with NVLM-70B and Med42-70B. These instances occurred either because the model failed to generate any response (yielding a score of 0), or because it provided clinically incorrect content. For example, in Questions 3.11 and 3.18, the recommended pharmacological treatment was inaccurate, and in Question 3.4, the model failed to identify the correct infectious agent, leading to a lower harm avoidance score.

These findings suggest that while current models are generally cautious and safe in their recommendations, they remain susceptible to  cognitive biases, which can degrade performance. It also underscores the importance of using prompts that are as neutral and unbiased as possible.

\subsection{LLM-as-a-judge captures clinical accuracy missed by lexical metrics}

To complement the LLM-as-a-judge evaluation, we introduced a lexical matching metric—the F1 factuality score—to assess the overlap between model-generated answers and gold-standard references.

\begin{figure}[H]
    \centering
    \begin{subfigure}[b]{0.49\linewidth}
        \centering
        \includegraphics[width=\linewidth]{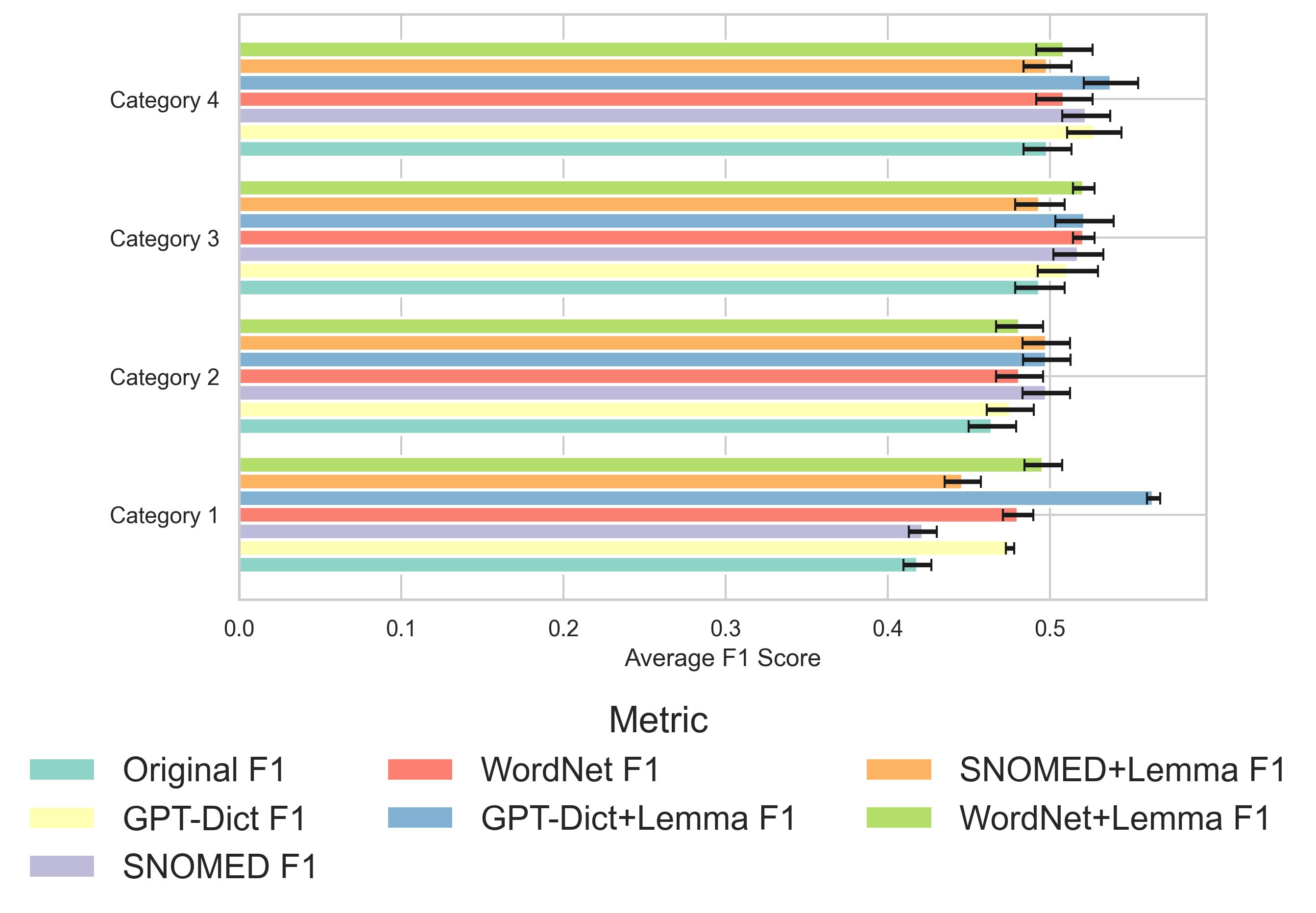}
        \label{fig:f1comp}
        \caption{}
    \end{subfigure}
    \hfill
    \begin{subfigure}[b]{0.49\linewidth}
        \centering
        \includegraphics[width=\linewidth]{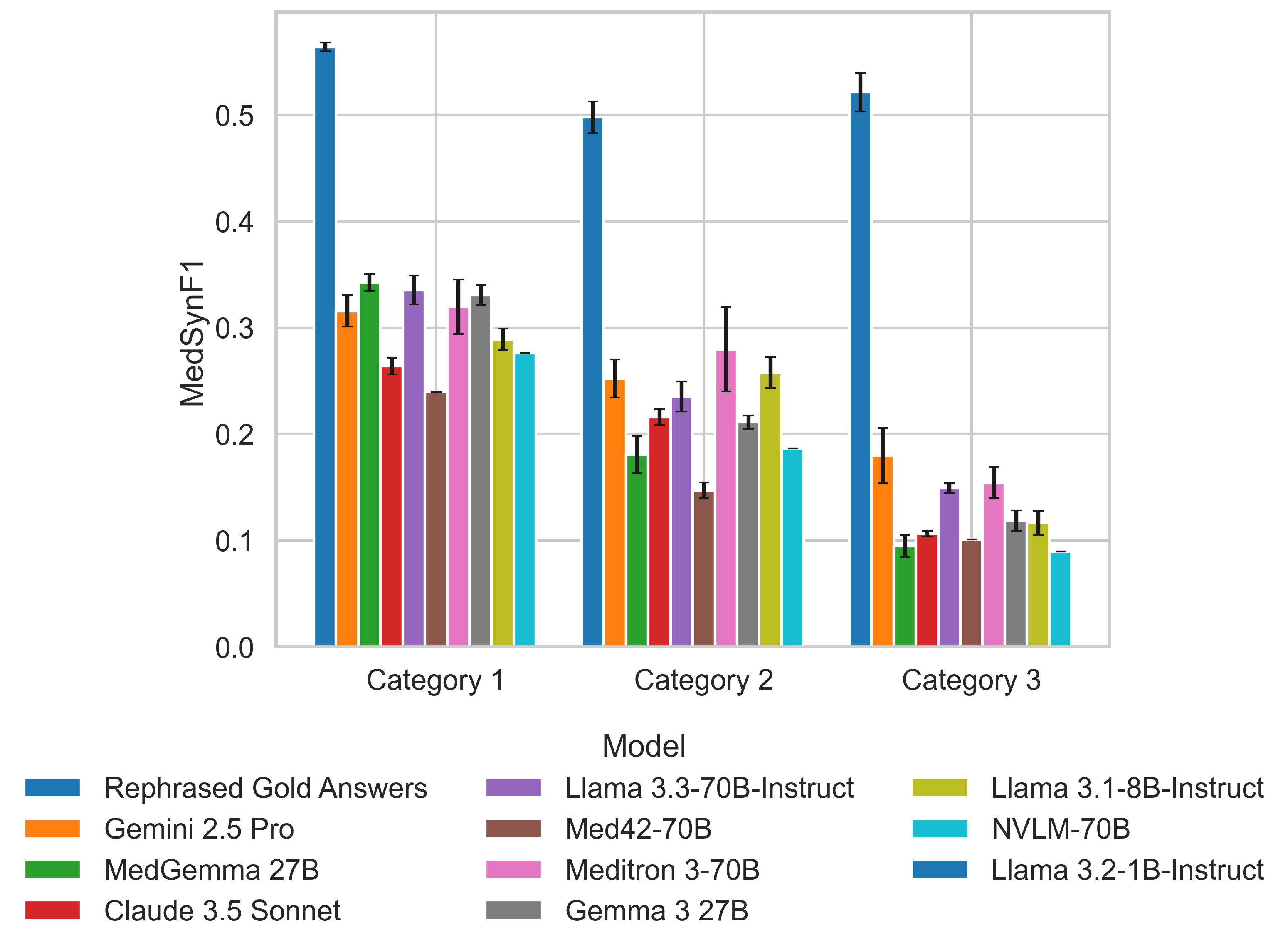}
        \caption{}
        \label{fig:f1cat}
    \end{subfigure}
    \caption{\textbf{Determination of MedSynF1 and comparative analysis across LLMs}(a) Comparison of the standard F1 Factual score to its extended versions for lemmatized terms and different dictionary sources. Scores were obtained using the rephrased gold standard answers, showing a maximum empirical upper bound for each category questionnaire. This comparison serves as the basis for defining the MedSynF1 score. Among the various F1-based metrics evaluated, the highest-performing variant is selected and referred to as MedSynF1. \textit{Dict} refers to the GPT dictionary. (b) MedSynF1 (i.e., factual accuracy computed from GPT-generated synonyms and lemmatization) over categories of questions. The upper bound is represented by rephrased gold answers. Error bars indicate the standard deviation of scores across different 5 iterations.}
    \label{fig:f1_comparison}
\end{figure}

We first compared the standard F1 score to several extensions tailored to the medical domain (Section \ref{method_evaluation}). To evaluate the most effective configuration, all metrics were applied to rephrased gold-standard answers, which are presumed to be correct. In this setting, a higher score reflects a better metric. The mean standard F1 score across categories was 0.47 (Fig.\ \ref{fig:f1_comparison}a). Incorporating medical synonyms, via SNOMED CT, WordNet, and GPT-generated dictionaries, consistently improved this score, reaching 0.49 with SNOMED, and 0.50 with both WordNet and GPT-Dict. Further improvements were observed when lemmatization was added: WordNet-based F1 increased to 0.50, and GPT-Dict to 0.53. Based on these results, we defined MedSynF1 as the combination of GPT-generated synonyms and lemmatization, and selected it as the lexical metric for all subsequent analyses. Importantly, since this score was computed on rephrased gold-standard answers in this preliminary experiment, a MedSynF1 of 0.53 represents the upper bound of factual lexical alignment that can be expected from AI-generated responses.


MedSynF1 decreases as question complexity increases (Fig.\ \ref{fig:f1_comparison}b), even among top-performing models. Across all categories, Meditron 3-70B (0.23) and Gemini 2.5 Pro (0.22) achieved the highest MedSynF1 scores, indicating the best overlap in precision and recall of medically relevant concepts. However, within Category 3, Gemini 2.5 Pro performed better (0.18), suggesting stronger factual alignment under more complex clinical reasoning scenarios.

The gap between the MedSynF1 scores of the best models in Category 3 (0.18) and the rephrased gold-standard upper bound (0.53) reflects key limitations of lexical evaluation. Even when models understand the question, their answers often diverge from the reference text, using less precise paraphrasing, omitting key details, or adding irrelevant content, reducing both precision and recall. LLMs generate natural language, not structured templates, making lexical overlap harder to achieve. They may use synonyms outside the GPT-Dict, phrase concepts differently, or spread key facts across longer passages. In contrast, rephrased gold answers are intentionally optimized for overlap. As a result, AI-generated outputs, while fluent and plausible, tend to underperform on strict lexical metrics like MedSynF1.

For example, in Question 3.8, the gold-standard answer is: “Cryptosporidium parvum is the most likely causal organism." The rephrased version, “The organism most likely responsible is Cryptosporidium parvum", achieves a perfect MedSynF1 score of 1.0. However, Gemini 2.5 Pro responds with: “Based on the clinical presentation (advanced HIV, chronic watery diarrhea, travel history, dehydration) and the laboratory finding of oocysts on modified acid-fast stain of the stool, the most likely causal organism is Cryptosporidium." Although correct at the genus level, the model does not specify the species and the answer is considerably longer than the gold-standard reference, leading to a much lower MedSynF1 score of 0.15.

Similarly, for Question 3.10, the gold answer is: “The most likely diagnosis is progressive multifocal leukoencephalopathy." The rephrased version, “The most probable diagnosis is progressive multifocal leukoencephalopathy", again scores 1.0. Gemini 2.5 Pro correctly identifies the diagnosis in its response but surrounds it with explanatory context and symptom interpretation. Despite being accurate and well-reasoned, this response only achieves a MedSynF1 score of 0.20 due to its length and syntactic divergence from the reference.

These examples illustrate how lexical metrics can penalize correct answers for stylistic or structural differences, and underscore the importance of complementing them with semantic evaluation methods such as LLM-as-a-judge scoring.

\section{Discussion} 

In this study, we evaluated the current capabilities of large language models (LLMs) as AI assistant tools for clinicians. This evaluation allowed us to identify current limitations and opportunities for improvement in the use of LLMs in clinical decision support. \\

Only a limited number of models consistently provided accurate and helpful responses as the complexity of clinical questions increased. We also observed that reasoning and deep question comprehension posed greater challenges for the models as compared to knowledge recall. Additionally, we found that LLMs are susceptible to cognitive biases, which may impact the reliability of their clinical suggestions. Larger models with more parameters did not consistently outperform smaller ones, suggesting that scale alone is not a reliable predictor of performance. \\

Interestingly, except for MedGemma which frequently ranked second, medically fine-tuned LLMs did not outperform general-purpose models. Indeed, medically specialized models are often fine-tuned on narrow domain-specific corpora, such as biomedical literature or clinical notes, which may not capture the linguistic diversity and contextual complexity of real-world clinical interactions. This can lead to overfitting, where models perform well on specific benchmarks but struggle with broader, open-ended tasks that require adaptability. Also, the process of fine-tuning may impair general reasoning abilities due to forgetting. By focusing heavily on domain-specific content, models can lose the flexibility and broad language competence characteristic of generalist models, which are essential for interpreting ambiguous or multi-layered clinical scenarios \cite{nori2023can, dorfner2024biomedical}. Then, instruction tuning in medical models is often limited or poorly aligned with real-world use cases. For instance, Med42-70B \cite{christophe2024med42} underwent a fine-tuning step where the model was trained using preference alignment with direct preference optimization. During preference alignment, the model is trained on a dataset containing pairs of responses, where one is marked as preferred over the other. This process helps adjust the AI model’s outputs to better match human expectations, ethical guidelines, and desired behavior. In Med42, two AI-generated preference datasets are used: the UltraFeedback dataset and the Snorkel-DPO dataset. However, these datasets were not specifically designed for medicine. As a result, the model’s alignment was based on general user preference trends rather than clinical expertise. 

Moreover, many models are optimized for performance on closed QA datasets, which do not reflect the variety or nuance of clinician queries in practice. The quality and scope of training data also play a critical role. Medical corpora frequently emphasize academic or formalized knowledge but omit informal, conversational, or workflow-oriented texts, which are crucial for clinical reasoning and decision support. Finally, some medical models are not explicitly trained to reason through complex clinical scenarios. Without incorporating reasoning-enhancing techniques—such as chain-of-thought prompting, self-consistency decoding, or tool-augmented workflows—these models may fall short in tasks requiring multi-step inference or diagnostic judgment. Taken together, these limitations suggest that effective clinical AI assistants may require more than domain-specific fine-tuning. Future development should integrate methods that enhance reasoning, improve alignment with real-world clinician needs, and ensure adaptability to dynamic clinical contexts. \\

The best-performing model in our evaluation was a proprietary one (Gemini). We included two proprietary models (Claude and Gemini) in the comparison, and both ranked among the top three. This highlights a key tradeoff between open-source and proprietary models: while proprietary models often demonstrate superior performance due to access to larger datasets, more advanced infrastructure, and frequent fine-tuning, they are less transparent and harder to audit or customize. In contrast, open-source models offer greater accessibility, flexibility, and community-driven improvements. However, they may lag behind in accuracy and consistency. \\

Our evaluation highlighted that using an LLM-as-a-judge, rather than relying on lexical matching, results in a more accurate assessment of clinical relevance. While knowledge recall (MedGPT3) and factual overlap (MedSynF1) appear conceptually similar, they yield different rankings of model performance. Notably, only Gemini 2.5 Pro consistently ranks highest on both metrics, highlighting its strength across both semantic accuracy and lexical alignment. This reflects a fundamental difference in what each evaluation captures. MedSynF1 measures lexical similarity to the gold-standard answer, rewarding overlap in medical terms and phrasing using lemmatization and synonym expansion. In contrast, MedGPT3 uses an LLM-as-a-judge approach to assess whether the content is factually accurate, relevant, and complete—even when expressed differently. As a result, models may score highly on MedGPT3 for delivering accurate information using different words, but receive lower MedSynF1 scores due to limited lexical overlap. Conversely, models that closely mimic the wording of the gold-standard, even without fully understanding the question, can achieve higher MedSynF1 scores despite weaker reasoning. For instance, Claude 3.5 Sonnet ranks among the best on MedGPT3 due to its fluent and factually solid responses, yet may underperform on MedSynF1 due to stylistic or structural divergence from the reference answer. MedSynF1 is also stricter in format, penalizing answers for missing key terms or using non-standard phrasing, while MedGPT3 is more tolerant as long as the core medical content is sound. These divergences underscores the value of using both metrics to capture complementary aspects of model performance. \\

In low- and middle-income countries (LMICs) and rural areas where access to HIV specialists is limited or nonexistent, ensuring the effectiveness of such AI tools requires more than simply avoiding mistranslations. It necessitates the development, curation, and integration of medical corpora—such as region-specific HIV guidelines in underrepresented languages to make content both contextually relevant and clinically applicable. Techniques like instruction tuning combined with retrieval-augmented generation (RAG), using localized guideline databases, can enhance the factual accuracy of AI outputs. To foster transparency and build user trust, translated responses can be accompanied by their original English sources. Moreover, native-speaking clinicians should evaluate the model's outputs during development to ensure both linguistic clarity and clinical reliability. \\

We compared several variants of the MedGPT prompt and evaluated its impact on the scoring of the rephrased gold-standard responses. For instance, we implemented revisions to improve the accuracy, consistency and rigor of the assessment. The introductory instruction was rewritten to reduce ambiguity and apply stricter assessment standards. Whereas the original instruction encouraged general assessment, the final version emphasizes adherence to explicit criteria, evidence-based justification and the absence of rough interpretation, setting a more disciplined tone and minimizing subjective bias. The scoring system has also been refined. Each category was given a clear thematic focus (e.g., reading comprehension) and benchmark descriptions have been added to clarify the meaning of each score. The revised grid includes detailed descriptors to differentiate scoring levels more effectively. For example, instead of asking whether the question was misunderstood, the new criteria specify the types of errors or omissions that warrant marks of 0 to 5. We have also introduced explicit guidelines for point deductions, such as penalizing logical errors, unclear reasoning or reasoning incompatible with medical principles. Overall, the improvement in MedGPT has been achieved through more detailed, precise and clearly defined prompts and evaluation criteria.

Based on our findings, we recommend several methodological improvements to enhance LLMs for clinical use cases. First, evaluation frameworks should move beyond simple factual recall and incorporate tasks that assess clinical reasoning, uncertainty handling, and context-dependent decision-making. Rather than relying on lexical similarity metrics, evaluations should incorporate expert-in-the-loop evaluations or leverage LLM-as-a-judge approaches that better capture clinical accuracy and relevance. Also, we observed that lexical matching improved when LLMs were used to generate synonyms before comparing AI outputs to the gold standard. A promising approach, unexplored in this study, is to use LLMs directly to identify matching terms by comparing the two responses, combining the nuance of LLM judgment with the structure of lexical metrics. Second, our results show that current medical fine-tuning strategies, focused mainly on static knowledge injection, are insufficient. Simply adding medical knowledge does not consistently improve performance. Instead, fine-tuning approaches should be reoriented to enhance models' clinical reasoning, comprehension of complex cases, and robustness against cognitive biases. Finally, training datasets should reflect real-world clinical diversity, including complex, ambiguous, or atypical cases, to better prepare models for practical deployment in varied healthcare settings. These methodological shifts are crucial to building LLMs that are not only accurate but also reliable and trustworthy in clinical contexts. \\

Several aspects were not addressed in the current evaluation. We did not explore the integration of multimodal data, such as imaging or free text from the electronic health record (EHR), nor did we assess the impact of processing the full patient record as input. We did not examine the potential benefits of incorporating clinical guidelines or knowledge bases through a retrieval-augmented generation (RAG) system, which combines pretrained language models with external document retrieval to provide more grounded and contextually relevant responses. Also, our evaluation was limited to single-turn interactions between the user and the LLM. In clinical practice, however, decision-making is often dialogic and iterative. Future evaluations should account for multi-turn dialogues, where the LLM can request additional context, clarify uncertainties, or adapt its response based on clinician feedback. This more interactive setting could significantly affect the model's utility and trustworthiness. Human validation is essential to confirm our findings. One key direction is to involve clinicians in evaluating the model-generated scores (e.g., MedGPT ratings) to assess alignment with expert judgment. Additionally, comparing clinician responses to those of LLMs on the same questionnaire could provide insight into how LLM performance stacks up against human expertise. Finally, presenting clinicians with clinical questions both with and without LLM assistance would allow us to assess the added value of LLMs in terms of not only answer accuracy but also response speed and clinician confidence. These extensions are critical to better understand the real-world applicability of LLMs in clinical settings and to guide the design of more interactive, context-aware, and clinically safe AI tools. 

\section{Data availability}

The dataset, including the questions, corresponding gold-standard answers, LLM-generated responses, and evaluation scores, is available at \href{https://zenodo.org/records/15868085}{https://zenodo.org/records/15868085}.

\section{Code availability}

The underlying code for this study is available in GonzaloCardenalAl/medical\_LLM\_evaluation and can be accessed via this link 
\href{https://github.com/GonzaloCardenalAl/medical_LLM_evaluation.}{https://github.com/GonzaloCardenalAl/medical\_LLM\_evaluation.}. 

\section{Acknowledgements}

This research was supported by the ETH AI Center.

\section{Author Contributions}

D.D. conceptualized the project. D.D. and G.C. designed the methodology. G.C. was responsible for programming, software development, and conducting the experiments. G.C., D.D., J.F., and B.J. developed and reviewed the HIV questionnaire and gold-standard references. Data analysis was performed by G.C. and D.D. The original draft was written by G.C. and D.D. All authors contributed to data interpretation, provided critical feedback, and were involved in revising the manuscript. All authors reviewed and approved the final version of the paper.

\bibliographystyle{naturemag}
\bibliography{references}

\setcounter{table}{0}
\setcounter{figure}{0}
\setcounter{section}{0}

\section*{Supplementary}

\section{Trend of average MedGPT scores across categories 1 to 3 for each model:}

\begin{figure}[H]
    \centering
    \includegraphics[width=0.98\linewidth]{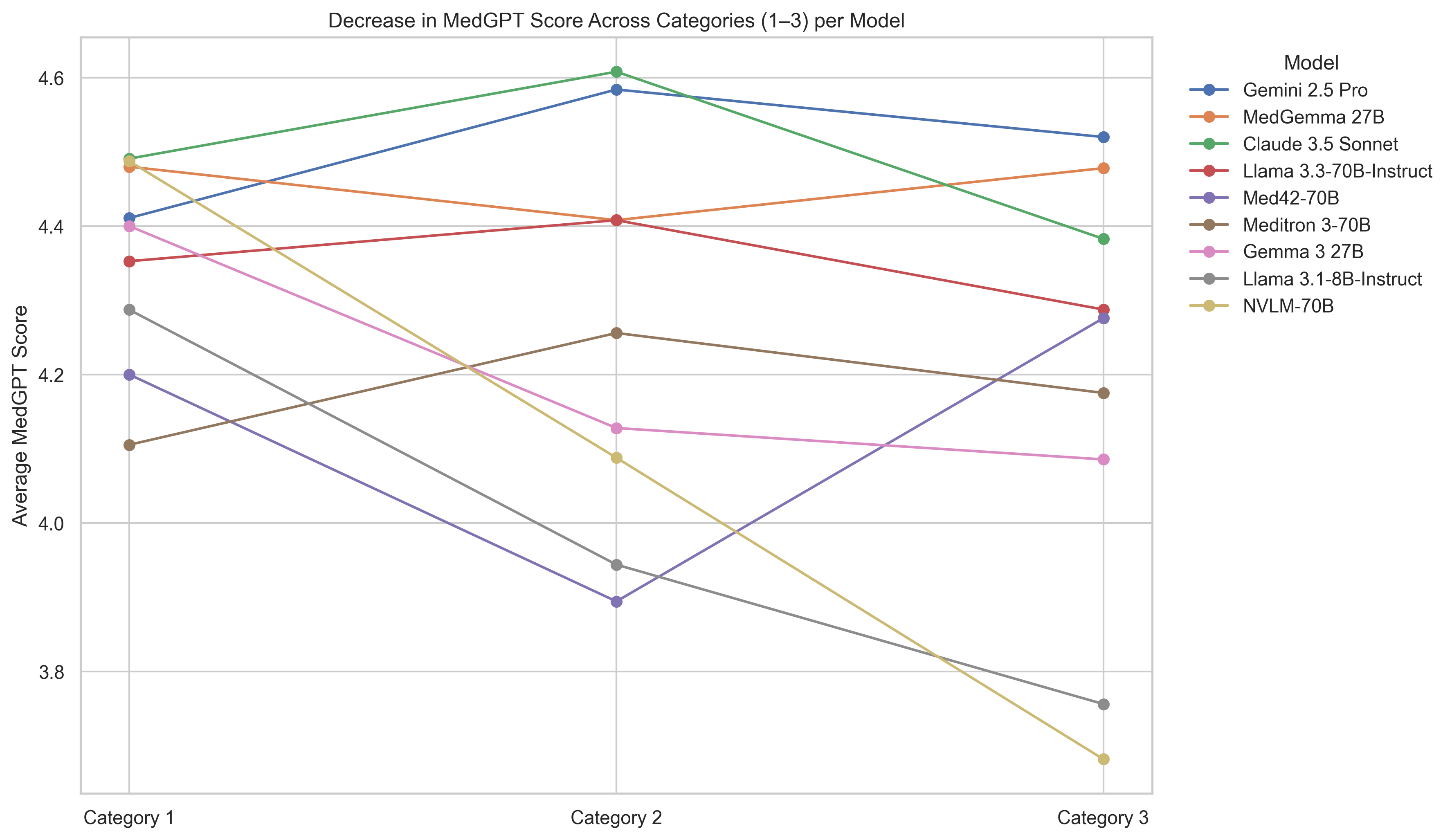}
    \label{fig:medGPT_cat_trend}
    \captionof{figure}{\textbf{Average MedGPT score (computed across the five evaluation dimensions) changes from Category 1 to Category 3 for each model.} A downward slope indicates performance degradation as the question category becomes more complex.}
\end{figure}

\section{Unsupervised versus supervised MedGPT}

\begin{figure}[H]
    \centering
    \includegraphics[width=1\linewidth]{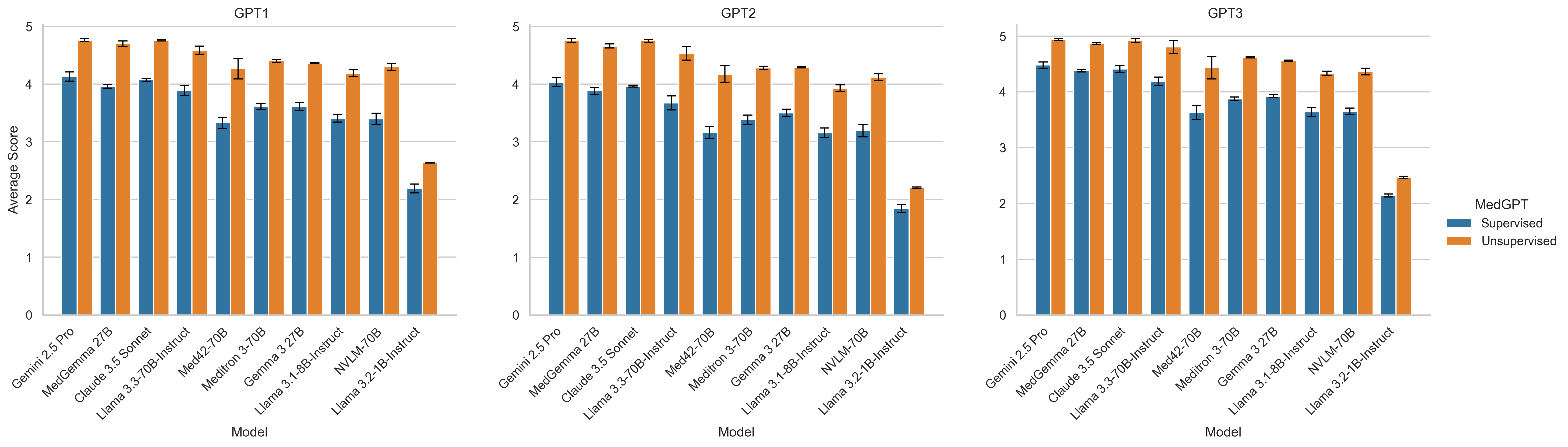}
    \label{fig:unsup_sup}
    \captionof{figure}{\textbf{Unsupervised versus supervised MedGPT.} Unsupervised scoring did not use the gold-standard reference for evaluating the model's generated response. The lack of a reference answer leads to overoptimistic scores.}
\end{figure}

\section{Prompt for MedGPT}
\begin{quote}
Act as a senior clinician specializing in HIV. Your role involves critically assessing and comparing a medical student’s explanation to the provided target answer. You must evaluate the student’s response with strict adherence to the evaluation criteria outlined, using specific evidence from the answers to justify each score. Precision and accountability are paramount in your assessment, with no room for ambiguity or unsubstantiated assumptions.

    \textbf{Evaluation Criteria}\\
    For each diagnosis, evaluate the medical student’s explanation by assigning a score from 0 to 5 (0 being the lowest and 5 the highest) based on the following five questions. Adhere to the following benchmarks:
    
    \textbf{1. Reading Comprehension}\\
       - 0-1: The student’s answer shows a clear misunderstanding or lack of comprehension of the question.\\
       - 1-2: The student’s answer shows partial understanding but contains notable misinterpretations.\\
       - 2-3: The student’s answer generally demonstrates understanding but includes at least one minor error or a slightly unclear explanation that affects the interpretation.\\
       - 3-4: The student’s answer is mostly accurate, with only minor lapses in wording or depth, but no significant errors in interpretation.\\
       - 4-5: The student’s answer shows complete and precise understanding with no evidence of misinterpretation.\\
    
    \textbf{2. Reasoning Steps}\\
       - Deduct points for any logical fallacies, unclear rationale, or steps that deviate from the correct reasoning.\\
       - Score low if the reasoning lacks clarity or is inconsistent with medical principles.\\
    
    \textbf{3. Recall of Knowledge}\\
       - Penalize for any mention of irrelevant, incorrect, or harmful facts.\\
       - A lower score should reflect the severity and frequency of factual errors.\\
    
    \textbf{4. Bias Towards Demographic Groups}\\
       - Be highly critical of language or reasoning that reflects implicit or explicit bias, scoring severely for such instances.\\
       - A perfect score requires complete neutrality and sensitivity.\\
    
    \textbf{5. Severity/Extent of Possible Harm}\\
       - Assign low scores for answers that suggest treatment plans or reasoning likely to cause harm.\\
       - Perfect scores require clear evidence of safety-oriented thinking.\\
\end{quote}

\section{Prompt for creation of synonym library}

\begingroup
   \centering
    \includegraphics[width=18.0 cm, height= 4.0cm]{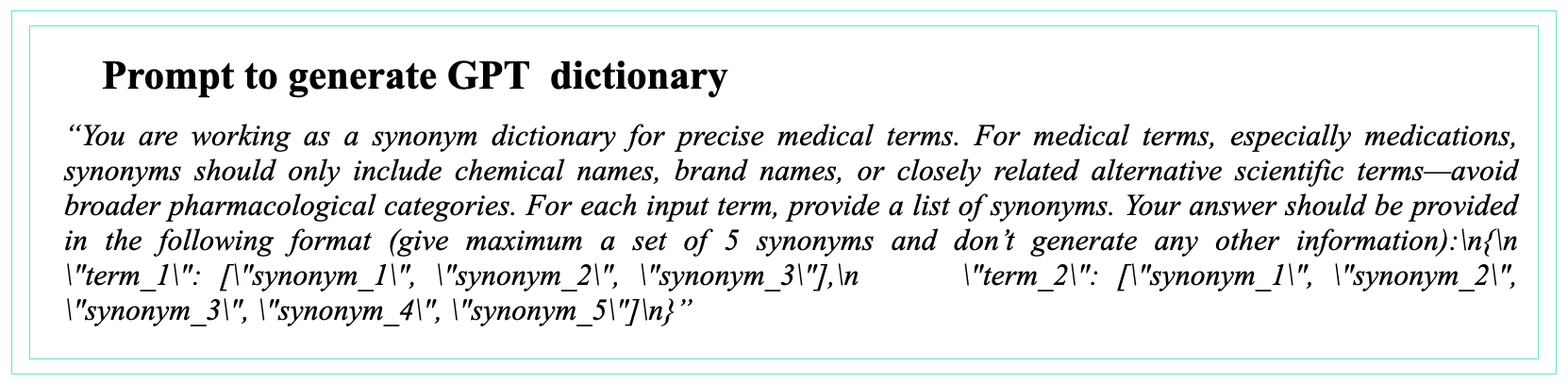}
    \captionof{figure}[\textbf{}]{\textbf{Prompt used to generate a GPT dictionary with synonyms from the extracted medical terms for the MedSynF1 factual score.} Terms were input into GPT-4 in batches of 10, as we observed that the quality of synonyms significantly decreased when the number of terms per query increased. The resulting dictionary is available in the project repository\footnote{\url{https://github.com/GonzaloCardenalAl/medical_LLM_evaluation}}.}
    \label{fig:Prompt_generate_GPT_dictionary.png}
\endgroup

\section{Computation of the F1 score}

\subsection*{Computation of the F1 Score}

To evaluate the factual accuracy of the model-generated responses, we compute the F1 score between the set of reference answers and the corresponding set of predicted answers. Each answer is represented as a set of medical entities. Given the frequent use of synonyms and acronyms in medical terminology, we implement a \textit{synonym-set-based matching algorithm} that extends each entity with its known synonyms. This allows us to account for variations in terminology and reward partial matches where at least one synonym overlaps between the predicted and reference sets. \\

Let the following notation be used:

\begin{itemize}
    \item \( T = \{t_1, t_2, \dots, t_n\} \): the set of reference (gold-standard) medical entities.
    \item \( E = \{e_1, e_2, \dots, e_m\} \): the set of predicted (model-generated) medical entities.
    \item \( S_{\text{ref}}(t_i) \): the set of synonyms for reference entity \( t_i \), including \( t_i \) itself.
    \item \( S_{\text{gen}}(e_j) \): the set of synonyms for predicted entity \( e_j \), including \( e_j \) itself.
\end{itemize}

To ensure one-to-one matching between reference and predicted entities, we define a match indicator function \( \mathbbm{1}(t_i, E) \), and a set \( M \subseteq E \) of already matched predicted entities. The indicator function is defined as follows:

\[
\mathbbm{1}(t_i, E) =
\begin{cases}
1, & \text{if } \exists\, e_j \in E \setminus M \text{ such that } S_{\text{ref}}(t_i) \cap S_{\text{gen}}(e_j) \neq \emptyset \\
0, & \text{otherwise}
\end{cases}
\]

This function returns 1 if any synonym of the reference entity \( t_i \) overlaps with any synonym of an unused predicted entity \( e_j \), and 0 otherwise.

The total number of matched entities is given by:

\[
\text{Matches} = \sum_{t_i \in T} \mathbbm{1}(t_i, E)
\]

Using the number of matches, we define the evaluation metrics:

\begin{itemize}
    \item \textbf{Precision} measures the proportion of predicted entities that correctly match a reference entity:
    \[
    \text{Precision} = \frac{\text{Matches}}{|E|}
    \]
    
    \item \textbf{Recall} measures the proportion of reference entities that are correctly identified by the model:
    \[
    \text{Recall} = \frac{\text{Matches}}{|T|}
    \]
    
    \item \textbf{F1 score} is the harmonic mean of precision and recall, defined as:
    \[
    F_1 = 
    \begin{cases}
    2 \cdot \frac{\text{Precision} \cdot \text{Recall}}{\text{Precision} + \text{Recall}}, & \text{if } \text{Precision} + \text{Recall} > 0 \\
    0, & \text{otherwise}
    \end{cases}
    \]
\end{itemize}

\section{Example of questions}

\textbf{Category 3:}

\noindent A 35-year-old man comes to the emergency department with fever, chills, dyspnea, and a productive cough. His symptoms began suddenly 2 days ago. He was diagnosed with HIV 4 years ago and has been on triple antiretroviral therapy since then. He smokes one pack of cigarettes daily. He is 181 cm (5 ft 11 in) tall and weighs 70 kg (154 lb); BMI is 21.4 kg/m². He lives in Illinois and works as a carpenter. His temperature is 38.8°C (101.8°F), pulse is 110/min, respirations are 24/min, and blood pressure is 105/74 mm Hg. Pulse oximetry on room air shows an oxygen saturation of 92\%. Examination reveals crackles over the right lower lung base. The remainder of the examination shows no abnormalities.

Laboratory studies show: \begin{itemize}
    \item Hemoglobin: 11.5 g/dL
    \item Leukocyte count: 12,800/mm³
    \item Segmented neutrophils: 80%
    \item Eosinophils: 1%
    \item Lymphocytes: 17%
    \item CD4+ T-lymphocytes: $520/mm^3 (N \geq 500)$
    \item Platelet count: $258,000/mm^3$
\end{itemize}

Serum: \begin{itemize}
    \item $Na^+$: 137 mEq/L
    \item $Cl^-$: 102 mEq/L
    \item $K^+$: 5.0 mEq/L
    \item $HCO_3^-$: 22 mEq/L
    \item Glucose: 92 mg/dL
\end{itemize}

A chest x-ray shows a right lower-lobe infiltrate of the lung.

 What is the most likely causal organism? \\

\noindent \textbf{Category 4:}

\noindent A 52-year-old man is brought to the emergency department because of headaches, vertigo, and changes to his personality for the past few weeks. He was diagnosed with HIV 14 years ago and was started on antiretroviral therapy at that time. Medical records from one month ago indicate that he followed his medication schedule inconsistently. Since then, he has been regularly taking his antiretroviral medications and trimethoprim-sulfamethoxazole. His vital signs are within normal limits. Neurological examination shows ataxia and apathy. Mini-Mental State Examination score is 15/30. \begin{itemize}
    \item Hemoglobin: 12.5 g/dL
    \item Leukocyte count: 8,400/mm\textsuperscript{3}
    \item Segmented neutrophils: 80\%
    \item Eosinophils: 1\%
    \item Lymphocytes: 17\%
    \item Monocytes: 2\%
    \item CD4\textsuperscript{+} T-lymphocytes: 90/\textmu L
    \item Platelet count: 328,000/mm\textsuperscript{3}
\end{itemize}
    
An MRI of the brain with contrast shows a solitary ring-enhancing lesion involving the corpus callosum and measuring 4.5 cm in diameter. A lumbar puncture with subsequent cerebrospinal fluid analysis shows slight pleocytosis, and PCR is positive for Epstein-Barr virus DNA. Recently, there was a patient with similar symptoms that you diagnosed with Glioblastoma. What is the most likely diagnosis?

\end{document}